\documentclass{article}
\usepackage{iclr2023_conference}
\usepackage{amsmath, amsfonts,bm,graphicx}
\usepackage{zref}
\usepackage[toc,page, title, titletoc]{appendix}
\usepackage{hyperref}
\usepackage{times}
\usepackage{epsfig}
\usepackage{amssymb}
\usepackage{multirow, bigstrut}
\usepackage{listings}
\usepackage{arydshln}
\usepackage{xcolor}
\usepackage{tikz}
\usepackage{pgfplots}
\usepackage{comment}
\usepackage{tabu}
\usepackage{caption}
\usepackage{subcaption}
\def\x{{\mathbf x}}

\title{Mitigating climate and health impact of small-scale kiln industry using multi-spectral classifier and deep learning}

\author{Usman Nazir, Murtaza Taj \& Momin Uppal \thanks{Tackling Climate Change with Machine Learning workshop at ICLR 2023} \\
Lahore University of Management Sciences\\
\texttt{\{17030059, murtaza.taj, momin.uppal\}@lums.edu.pk}
\And
Sara Khalid \\
University of Oxford \\
\texttt{sara.khalid@ndorms.ox.ac.uk}
}

\iclrfinalcopy
\begin{document}
\maketitle

\vspace{-0.5cm}
\begin{abstract}
%% Extension of KilnNet paper 
Industrial air pollution has a direct health impact and is a major contributor to climate change. Small scale industries particularly bull-trench brick kilns are one of the key sources of air pollution in South Asia often creating hazardous levels of smog  that is injurious to human health. To mitigate the climate and health impact of the kiln industry, fine-grained kiln localization at different geographic locations is needed. Kiln localization using multi-spectral remote sensing data such as vegetation indices can result in a noisy estimates whereas relying solely on high-resolution imagery is infeasible due to cost and compute complexities. This paper proposes a fusion of spatio-temporal multi-spectral data with high-resolution imagery for detection of brick kilns within the ``Brick-Kiln-Belt'' of South Asia. We first perform classification using low-resolution spatio-temporal multi-spectral data from Sentinel-2 imagery by combining vegetation, burn, build up and moisture indices. Next, orientation aware object detector  YOLOv3 (with $\theta$ value) is implemented for removal of false detections and fine-grained localization. Our proposed technique, when compared with other benchmarks, results in a $21\times$ improvement in speed with comparable or higher accuracy when tested over multiple countries. 
%Furthermore, we combined the obtained kiln locations with heat signature values from Landsat-8 to identify illegal kiln activity during winter smog period and compared it with data from Environmental Protection Agency.

\end{abstract}
%
% \begin{keywords}
% Sentinel-2, Google Earth Engine, Spectral properties, Brick Kiln, YOLOv3\end{keywords}
\vspace{-0.25cm}
\section{Path to Climate Impact}
Industrial air pollution has a direct health impact and is a major contributor to climate change. Unregulated small-scale informal industries spread across vast areas are common in resource-limited settings and can be difficult to locate and monitor. Remote identification of kilns and monitoring of their carbon production can assist air pollution surveillance, regulation, and climate mitigation efforts. The exact numbering and location of kilns is needed to understand and address the kiln sector's climate and health impacts.
\vspace{-0.25cm}
\section{Introduction}
Vehicles and industries are considered as one of the major contributors of pollution resulting in low air quality and smog~\cite{haque2022impact}. According to an estimate $20\%$ of global black carbon emission is from the brick kilns~\cite{KilnFacts}. These kilns are also a major source of employment, however according to the Global Slavery Index of $2019$, the ``Brick-Kiln-Belt'' of South Asia (stretching between Afghanistan, Pakistan, India, and Nepal) is home to approximately $60\%$ ($24.3$ million individuals) of modern-day slavery~\cite{landman2019globalization}. Keeping in view the UN's Sustainable Development Goals (SDGs) $3.9$ and $8.7$ which specifically aim to address air pollution and forced labor respectively, mapping brick kilns in the South Asia region is an essential first step. 

Manual surveying of ``Brick-Kiln-Belt'' is infeasible due to the large geographical spread ($1,551,997~{km}^2$) as well as geographic boundaries. Due to recent advancements in machine learning as well as availability of remote sensing data, automated surveys are now more commonly used for such large-scale analysis~\cite{redmon2018yolov3, he2017mask, li2018h, xie2016transfer, you2017deep, cotrufo2018building}.
% The problem of mapping brick kilns in limited regions of the ``Brick-Kiln-Belt'' of South Asia has been addressed in the past, albeit to a limited extent, using both low-resolution~\cite{misrabrick, misra2020mapping} and high-resolution satellite imagery~\cite{nazir2020kiln, boyd2018slavery}. Some of these solutions are based on manual surveys~\cite{boyd2018slavery} in which a crowd-sourced methodology is utilized for manually annotating brick kilns in remote satellite imagery. On the other hand, while deep convolutional networks (CNN) have enormous potential for detection of objects in satellite imagery, their use requires a significant amount of training data which is non-existent for most of the cities in the region of interest. The existing solutions for automatically detecting brick kilns (such as those based on Faster R-CNN~\cite{foody2019earth} or Normalized Difference Vegetation Index (NDVI)~\cite{misrabrick}) are limited in their spatial coverage~\cite{nazir2020kiln} due to lack of annotated data. In addition, these methods lack generalization and scalability which is essential for the large-scale problem at hand. 
% Para on Brick Kiln using Remote Sensing
Recently, remote sensing images have also been used to analyze the extent of modern slavery~\cite{boyd2018slavery, jackson2018analysing, misrabrick, foody2019earth}. The ``Slavery from Space'' project~\cite{boyd2018slavery} proposed a crowd-sourced procedure to manually detect brick kilns from satellite imagery. They randomly sampled the potential kiln areas into $320$ cells of $100$ $\text{km}^2$ each. However, they were only able to manually annotate $30$ geographic cells (i.e. only $2\%$ of the entire Brick-Kiln-Belt). As a result, the manual crowd-sourced method lacks generalization and scalability as is evident from the resulting sparsely annotated maps. On the other hand, low-resolution multi-spectral satellite data has also been used to classify brick kilns in the region surrounding the Delhi state in India~\cite{misrabrick}. The analysis in this work was based on normalized difference vegetation index (NDVI) and transfer learning, which unfortunately is prone to generate a large number of false detections. In contrast, high-resolution satellite imagery has also been used to detect brick kilns to the east of Jaipur, which is the capital city of India's Rajasthan state~\cite{foody2019earth}. This work used Faster R-CNNs to automate the process of brick kiln identification in the given tile of images. However, owing to the large computational complexity, this approach is difficult to apply at a large scale. Moreover, it yields a high false positive rate for which the study proposed to train a two-staged R-CNN classifier to achieve acceptable performance which further increased the processing time. A more recent approach called KilnNet~\cite{nazir2020kiln} combined inexpensive an classifier with object detector in a two-stage strategy to address the issue of time complexity. This approach too is only based on high resolution satellite imagery and is infeasible due to data acquisition and processing costs.

% \begin{figure}
%     \centering
% \scalebox{0.8}{
%   \begin{tabular}{cc}
%             \includegraphics[width=.4\columnwidth]{SpectralProperties/challenges/kiln1.jpg} &
% 			\includegraphics[width=.4\columnwidth]{SpectralProperties/challenges/kiln1pixels.PNG} \\
% 			(a) & (b) \\
% 			\includegraphics[width=.4\columnwidth]{SpectralProperties/challenges/kiln2.jpg} &
% 			\includegraphics[width=.4\columnwidth]{SpectralProperties/challenges/kiln2pixels.PNG} \\
% 			(c) & (d) \\ 
%     \end{tabular}}
%     \caption{Exemplary brick kiln images as seen in the reference high resolution ($\frac{0.5~meter}{pixel}$) imagery (a, c) [Courtesy: Digital Globe, Google Earth] and the low resolution ($\frac{10~meters}{pixel}$) Sentinel-2 imagery (b, d) with Mean NDVI values with $Opacity = 1~\&~0.5$ [Courtesy: Google Earth Engine].}
%     \label{brickkilnpixels}
% \end{figure}

Here we also propose a two-stage strategy for automated detection of brick kilns. However, our approach is over $21\times$ faster than state-of-the-art (SOTA) benchmarks and mostly relies on freely available low resolution remote sensing data. Most existing object detection techniques in low resolution satellite imagery are significantly less accurate  whereas computation is very costly for high resolution satellite imagery. To overcome this our proposed methodology decouples classification and localization. Classification is performed using spectral properties while localization is accomplished using orientation adapted detector. This results in a coarse-to-fine search in which fine-grained orientation aware localization via object detection is performed as a second stage only on less than $10\%$ of the total region. This results in a $21\times$ improvement in speed in addition to improvement in accuracy. We tested our approach on three countries (Pakistan, India and Afghanistan) and showed that it is scalable as well as generalizable to varying structural, environmental and terrain conditions. 

\begin{figure}[t]
		\centering
        \includegraphics[scale=0.4, trim={0cm 0cm 0cm 0cm},clip]{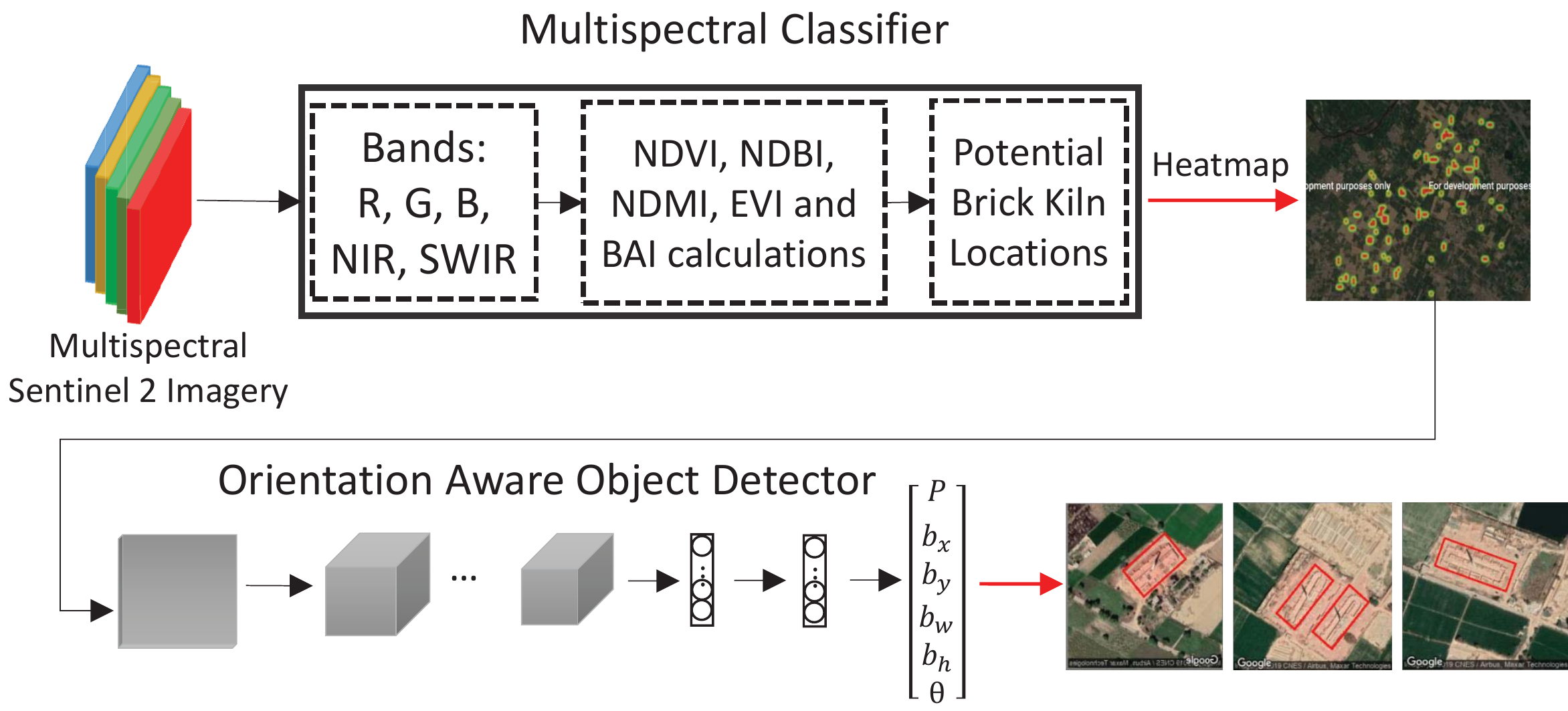}
		\caption{Illustrative example of working of the proposed approach. In the first step we apply a rule based classifier on spectral indices (NDVI, EVI, NDMI, NDBI, BAI) on regions of interest to classify brick kilns as shown in the heatmap (Satellite images courtesy Google Earth).}
		\label{illustrativeExample}
\end{figure}
% \begin{figure}[h]
% \scalebox{0.8}{
% 	\begin{tabular}{ccc}
% 		\multicolumn{3}{c}{\includegraphics[scale=0.35, trim={0cm 0cm 0cm 0cm},clip]{images/spectra.png}}\\
% 	\includegraphics[width=.3\columnwidth]{SpectralProperties/NDVI.jpg} 
% 	& \includegraphics[width=.3\columnwidth]{SpectralProperties/NDBI.jpg} &
% 	\includegraphics[width=.3\columnwidth] {SpectralProperties/NDMI.jpg} \\
% 		NDVI & NDBI & NDMI
% 	\end{tabular}}
% 	\caption{Reflectance Spectra for various targets and Spectral Indices of kiln locations of Punjab, Pakistan~\cite{bowker1985spectral}.}
% 	\label{spectra}
% \end{figure}

This paper has the following four technical contributions:
{\bf(i) Fusion of spectral Indices}: Classification is performed using mixture of spectral indices as shown by equation~\ref{fusionmi} in the paper.
{\bf(ii) Fusion of low resolution and high resolution imagery}: Our proposed approach processes input from low-resolution sensors for the generation of potential candidates for kilns which are then filtered via high-resolution input via Orietation-aware YOLOv3.
{\bf(iii) Processing large datasets}: The proposed strategy reduces the computational burden associated with processing of large datasets by fusion of low resolution and high resolution imagery. SOTA benchmarks take on average $674$ seconds to process three datasets in Table~\ref{evaluation} of the paper. Our proposed approach on the other hand reduces this compute time to $38$ seconds only.
{\bf(iv) Detection of other objects}: Our multi-sensor and multi-stage strategy can also be used to detect other objects that have differentiable spectral signatures and well defined shapes e.g. industrial units, oil tanks, warehouses, tennis courts, parking lots etc. Here we have only demonstrated its application for geo-localization of brick kilns. 
%Furthermore, provision of multi-sensor measurements occurs in many cases, such as  combination of thermal and optical camera, multispectral and optical camera, audio and video, aerial and street-view cameras, blood biomarkers and X-Rays, structured light and optical cameras and so on. Our idea can be extended for these sensor combinations as well. 

%The remainder of this paper is organized as follows. In Section~1, we provide an overview of the challenges associated with kiln detection from remote sensing data. This is followed by Section~2 which describes the proposed method based on spectral properties and location aware detector, while results and evaluation are provided in Section~3. Section~4 concludes the paper and also provides potential avenues of future research.
%
% ==================================================================
\vspace{-0.25cm}
\section{Proposed Methodology}
Brick kilns are typically identifiable through a visual inspection of satellite imagery. However, while considering a large geographic area, several inherent complexities in satellite imagery make automated detection of brick kilns a challenging task. This include, but are not limited to, i)  variations in imaging sensors, ii) differences in kiln structure across the countries, iii) dynamic kiln surroundings and iv) variations in luminosity, seasonal changes, and pollution levels etc. It is particularly challenging to identify brick kilns in low resolution imagery ($\frac{10-30~meters}{pixel}$). 
%as shown in Fig.~\ref{brickkilnpixels} (b, d). 
Existing pixel classification along with transfer learning approach~\cite{misrabrick, misra2020mapping} for detection of brick kilns is not scalable as well as generalizable at large scale due to cost complexities. In our proposed multi-spectral approach brick kilns are classified using spectral indices (without transfer learning) due to the fact that reflectance spectra of different land covers are different.  The indices are designed to exploit these differences to accentuate particular land cover types. The land covers are separable at one or more wavelengths due to spectral reﬂectance of different materials (see Fig.~\ref{fig:spectralIndices} and Fig.~\ref{brickkilnpixels} in Appendix~\ref{appA}).
%~\cite{yan2015urban}. 
Brick kilns being man-made structures have a high built-up index. Unlike other man-made structures and due to the specific nature of work, the kiln surrounding has a low vegetation index. Furthermore the baking process and smoke from chimneys also result in a high burn index. Thus in this work we classify brick kilns using mixture of spectral indices namely NDVI, EVI, NDBI, NDMI and BAI (see Appendix~\ref{appA}). The proposed architecture is shown in Fig.~\ref{illustrativeExample}.

\begin{comment}
\begin{figure}[t]
\scalebox{0.9}{
\begin{tabular}{cccc}
	\includegraphics[width=.22\columnwidth]{SpectralProperties/indices/satellite image.png}
	& \includegraphics[width=.22\columnwidth]{SpectralProperties/indices/NDVI.png}
	& \includegraphics[width=.22\columnwidth]{SpectralProperties/indices/moistureIndex.png}&
	\includegraphics[width=.22\columnwidth]  {SpectralProperties/indices/builtUp.png}
	\\
	Satellite Image & NDVI & NDMI & NDBI  
	\end{tabular}}
	\caption{Kiln surrounding has a low vegetation index (NDVI) and moisture Index (NDMI) and a high built-up index (NDBI); (Darker colour shows more index value).}
	\label{spectra}
\end{figure}
\end{comment}

%To tackle this issue we used a mixture of 5 spectral indices to differentiate brick kilns from other kind of chimneys.
% The unique spectral characteristics of brick kilns distinguish them from other classes such as built-up, vegetation and fallow-land even in low resolution imagery.  

\begin{table}[ht]
\caption{Table showing quantitative evaluation of the proposed approach compared with state-of-the-art methods. Top-3 ranking methods are in bold and, in particular, red (1st), violet (2nd) and black (3rd).}
\centering
\scalebox{0.56}{
	\begin{tabu}{c|c|ccccccccc}
		\hline
		 & \multicolumn{9}{c}{\bf{Classification Score}} \\ 
		\bf{Testing Datasets} & \bf{Network Architectures}& TP& TN & FP & FN & \bf{Duplicates} & \bf{Precision} & \bf{Recall}  & \bf{F1 score} & \bf{Time (seconds)} \\ \hline
		\multirow{5}{*} {Pakistan (Kasur)} 	
        & {Multi-spectral Approach} & 21 & - & 303 & 0 & 2 & 0.065 & 1 & 0.12 & \textcolor{red}{3} \\   
		& {{Two-Stage Faster R-CNN}} & 13 & 3090 & 1& 6& 0 & 0.93& 0.68& \textcolor{black}{\bf 0.79}& 195.5\\
		& {{Two-Stage SSD}} & 12 & 3090 & 1& 7& 0 & 0.92& 0.63& 0.75& 179.5\\
		& Kiln-Net (Two-Stage YOLOv3)~\cite{nazir2020kiln} & 19 & 3091 & 0& 0& 0 & 1 & 1& \textcolor{red}{\bf 1}& \textcolor{black}{\bf 162.5} \\ 
		& Proposed (Multispectral Two-Stage Strategy) &21 & - &1 &1 &2 & 0.95 & 1 &\textcolor{violet}{\bf 0.97} & \textcolor{violet}{\bf 7.04}
		\\ \cdashline{1-11}
		\multirow{5}{*} {India (New Delhi)}  	& \bf{Multi-spectral Approach} & 52 & - & 636 & 0 & 12 & 0.076 & 1 & 0.14 & \textcolor{red}{\bf{4}} \\
		& {{Two-Stage Faster R-CNN}} & 37 & 4441 & 1& 3& 0 & 0.97& 0.93& 0.95& 276.1\\
		& {{Two-Stage SSD}} & 38 & 4442 & 0& 2& 0 & 1& 0.95& \textcolor{black}{\bf 0.97}& 255.3\\
		& Kiln-Net (Two-Stage YOLOv3)~\cite{nazir2020kiln} & 40 & 4442 & 0& 0& 0 & 1 &  1 & \textcolor{red}{\bf{1}}& \textcolor{black}{\bf 232.8} \\
	& {\bf Proposed (Multispectral Two-Stage           Strategy)} & 52 & - &1 &0 &12 &0.98 & 1          & \textcolor{violet}{\bf 0.99}       &     
       \textcolor{violet}{\bf{8.16}}
	\\ \cdashline{1-11}
	\multirow{5}{*} {Afghanistan (Deh Sabz)} 
	& \bf{Multi-spectral Approach} & 406 & - & 1940 & 0 & 142 & 0.17 & 1 & 0.29 & \textcolor{red}{\bf{8}} \\
		& {{Two-Stage Faster R-CNN}} & 100 & 4097 & 5& 22& 0 & 0.95& 0.82& \textcolor{red}{\bf 0.88} & 553.2\\
		& {{Two-Stage SSD}} & 90 & 4094 & 8& 32& 0 & 0.92& 0.74 & \textcolor{black}{\bf 0.82}& 416.4\\
		& Kiln-Net (Two-Stage YOLOv3)~\cite{nazir2020kiln} & 85 & 4073 & 29& 37& 0 & 0.75 & 0.70 & 0.72 & \textcolor{black}{\bf 279.6}\\ 
		& {\bf Proposed (Multispectral Two-Stage Strategy)} & 198 & - & 17 & 66 & 142 & 0.92 & 0.75 & \textcolor{violet}{\bf{0.83}} & \textcolor{violet}{\bf{23.1}}
		\\ \cdashline{1-11}
	\end{tabu}}
	\label{evaluation}
\end{table}
\begin{comment}    
\begin{figure}[h]
\centering
\includegraphics[scale=0.25, trim={0cm 0cm 0cm 0cm},clip]{images/geeie.png}
\caption{Qualitative evaluation of our proposed Multi-spectral approach on region of Punjab, Pakistan. In first stage of our proposed two stage strategy, around $>99\%$ data is filtered out and only positive potential candidates (red pixels images) are passed to second stage for localization.  (Image courtesy Google Earth Engine).}
\label{geeie}
\end{figure} 
\end{comment}
\vspace{-0.25cm}
\subsection{Kiln Candidates via Multi-Spectral Classification}
\begin{figure}[t]
\centering
\scalebox{0.5}{ 
 \begin{tabu}{c|ccc}		
            %\multicolumn{3}{c}{\bf{Proposed Approach}}\\
            \hline
            Multi-spectral classification & & Orientation aware detector (Deep Learning) &\\
		\hline
            \multirow{2}[2]{*}[2mm]{\includegraphics[scale=0.30]{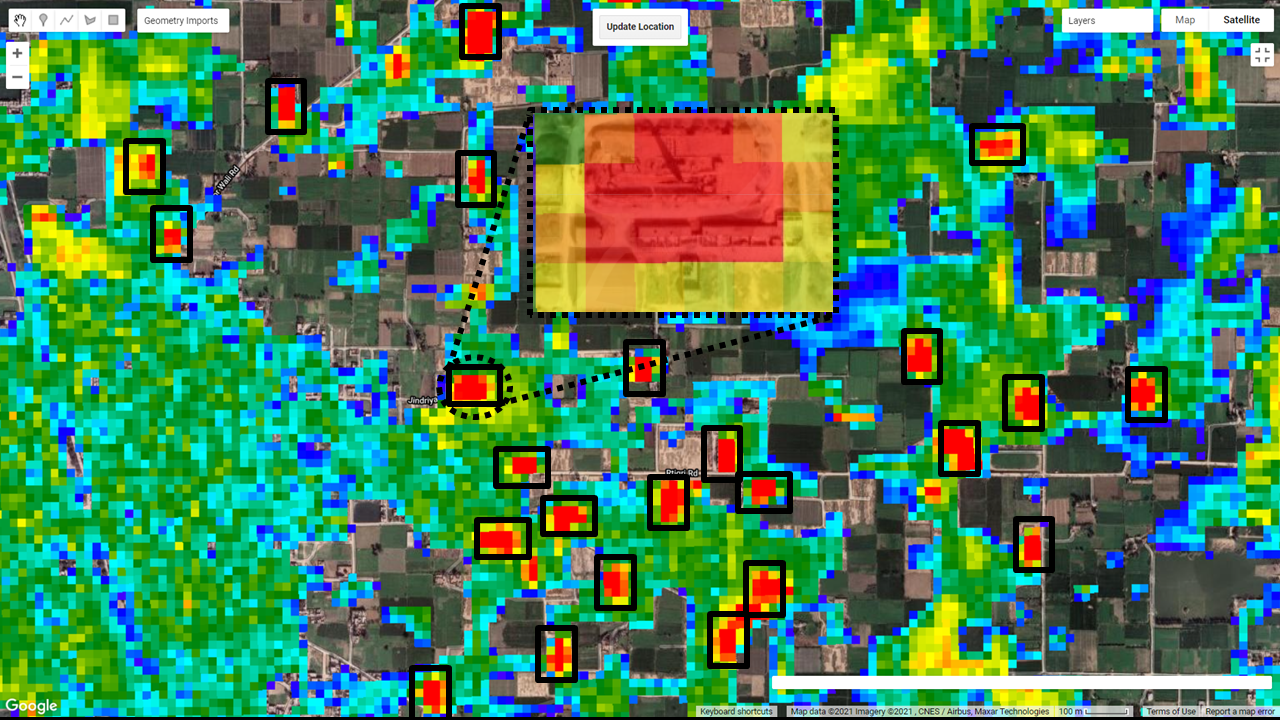}}
            &\includegraphics[width=.25\columnwidth]{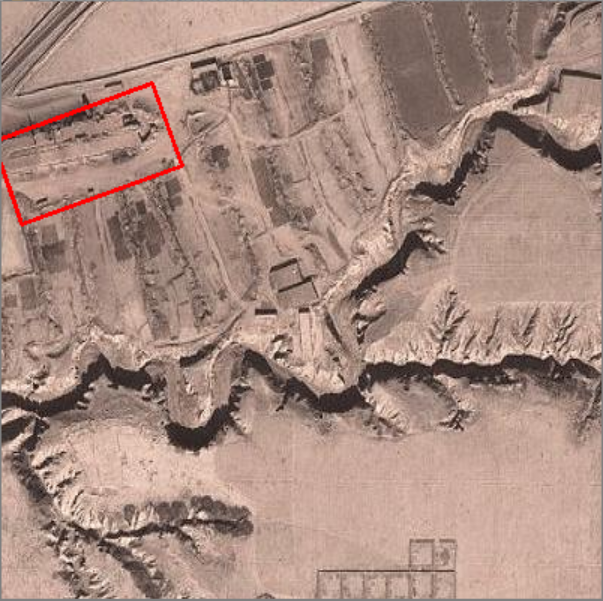} &\includegraphics[width=.25\columnwidth]{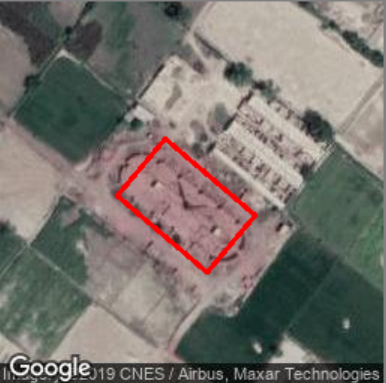}&
			\includegraphics[width=.25\columnwidth]{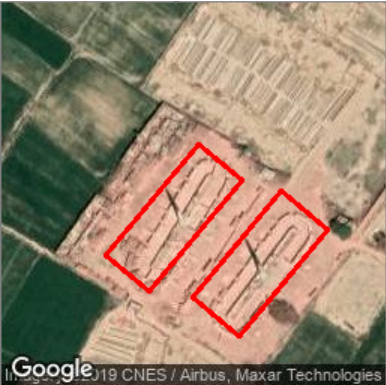} \bigstrut \\ 
			&\includegraphics[width=.25\columnwidth]{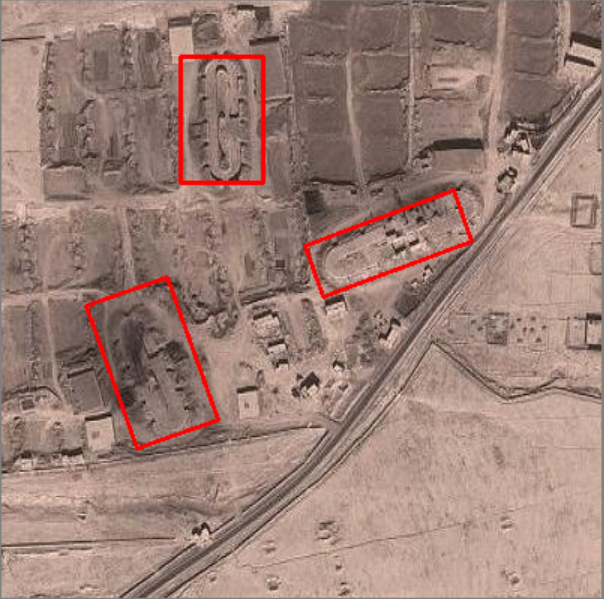} &\includegraphics[width=.25\columnwidth]{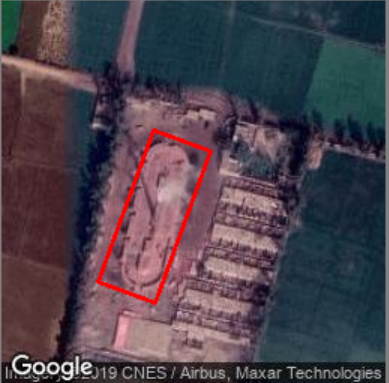}&
			\includegraphics[width=.25\columnwidth]{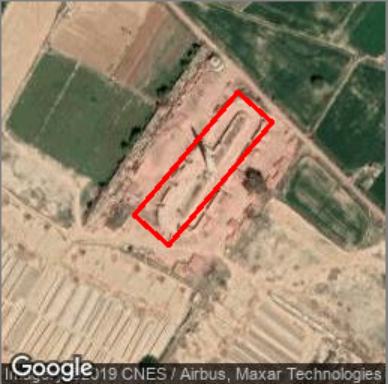}\bigstrut \\
	$99\%$ data is filtered out in classification stage & (a) Afghanistan & (b) Pakistan & (c) India  \\
        \hline
        (i) & &(ii)\\
			%\multicolumn{3}{c}{\bf{Dynamic Surroundings}}\\ \cdashline{1-3} \\	
		\end{tabu}}
	\caption{(i) Qualitative evaluation of our proposed Multi-spectral approach on region of Punjab, Pakistan. In first stage of our proposed two stage strategy, around $>99\%$ data is filtered out and only positive potential candidates (red pixels images) are passed to second stage for localization. (ii) Qualitative evaluation of Orientation aware YOLOv3. (Satellite images courtesy Google Earth).}
	\label{fig:Var}
\end{figure}

 Based on the assumption that kiln locations have low vegetation and moisture index whereas high build-up and burn area index, small or negative values of NDVI, EVI and NDMI with positive values of NDBI and BAI are classified as brick kilns (see Fig.~\ref{fig:Var} (i)). Thus our classification rule is defined as:
\begin{equation}\label{fusionmi}
\scriptsize		
f(x,y) =
  \begin{cases}
    1 \quad& if~~\text{$NDVI<0.2~\&~EVI<0.2$}~\&~\text{$NDMI<0~\&~NDBI>0$}~\&~\text{$ BAI > 5e^{-8}$}  \\   
    0 & \text{otherwise}
  \end{cases}
\end{equation}
where $(x,y)$ is a location in latitude and longitude and function $f(\cdot)$ gives the classification decision. We set the threshold of $0.2$ for NDVI and EVI as values $>0.2$ are considered as healthy vegetation~\cite{huete2002overview}. In the second stage we apply orientation aware detector (YOLOv3 with $\theta$ value) for false removal and kilns bounding boxes.

\vspace{-0.25cm}
\subsection{Orientation aware detector: YOLOv3}
%such as built-up, vegetation and fallow-land in low resolution imagery
Although the unique spectral characteristics of brick kilns distinguish them from other classes, however they are still confused with other small industries' chimneys as they exhibits similar spectral properties particularly NDVI and NDBI. We eliminated the resulting false detection via object detector. Unlike urban housing, kilns in South Asia are usually build at sparse locations which are mostly surrounded by agriculture land. Consequently they are usually built at arbitrary orientations. Thus the detection of axis aligned kilns~\cite{misra2020mapping, nazir2020kiln} is not applicable and results in increased missed detections. To address this problem, bounding box with orientation can be used~\cite{zhang2019r}. We therefore modified the YOLOv3 detector and added the neuron for regressing orientation with each bounding box. We only provided filtered images (potential candidates for kilns), obtained after classification stage, to the orientation-aware YOLOv3 model and obtained brick kiln bounding boxes as output results.

\vspace{-0.25cm}
\section{Quatitative and Qualitative Evaluation}
For detailed experimentation of our proposed procedure, we choose an evaluation dataset of three cities named Deh Sabz, New Delhi and Kasur from three different South Asian countries namely Afghanistan, India and Pakistan. For fair comparison we use the same geographical regions in these cities as defined in KilnNet paper~\cite{nazir2020kiln}. In addition, in order to generate training data for our object detector, we also manually annotated bounding boxes for each one of the $1300$ brick kiln images from the `Asia14' dataset~\cite{nazir2020kiln}.
%

% \pgfplotsset{compat=newest}
% \usetikzlibrary{decorations.pathmorphing}
% \begin{figure}
% \begin{tikzpicture}
% \begin{axis}[
%     every axis plot post/.style={/pgf/number format/fixed},
%     ybar=10pt,
%     bar width=12pt,
%     x=3cm,
%     ymin=0,
%     axis on top,
%     ymax=16,
%     xtick=data,
%     enlarge x limits=0.2,
%     legend style={at={(0.48,0.8)}},
%     %legend style={at={(0.02,0.02)},anchor=east},
%     symbolic x coords={Kasur,New Delhi, Deh Sabz},
%     restrict y to domain*=0:18, % Cut values off at 14
%     visualization depends on=rawy\as\rawy, % Save the unclipped values
%     after end axis/.code={ % Draw line indicating break
%             \draw [ultra thick, white, decoration={snake, amplitude=1pt}, decorate] (rel axis cs:0,1.05) -- (rel axis cs:1,1.05);
%         },
%     nodes near coords={%
%             \pgfmathprintnumber{\rawy}% Print unclipped values
%         },
%     axis lines*=left,
%     clip=false
%     ]
%  \addplot+[error bars/.cd,y dir=both,y explicit] coordinates {(Kasur,4.04)+- (0.0, 0.1) (New Delhi,4.16)+- (0.0, 0.13) (Deh Sabz,15.1)+- (0.0, 0.1)};
%  \addplot coordinates {(Kasur,162.5) (New Delhi,232.8) (Deh Sabz,279.6)};
% \legend{Proposed, KilnNet}
% \end{axis}
% \end{tikzpicture}  
% \captionof{figure}{Compute Time of Proposed Multispectral 2 stage strategy vs. KilnNet on Testing Regions: Kasur (Pakistan), New Delhi (India) and Deh Sabz (Afghanistan).}
% %\caption{}
% \label{stdgraph}
% \end{figure}
%
% =========================================================

%
We evaluated our proposed multispectral two-stage strategy with comparison to ResNet-152~\cite{he2016deep} classifier followed by the SOTA detector: Faster R-CNN~\cite{ren2015faster}, SSD~\cite{liu2016ssd} and YOLOv3~\cite{redmon2018yolov3}. A coarse-to-fine strategy is proposed that aims to filter the bulk of the data using spectral properties while the detector is only applied on a small amount of positive detections to generate localization information while filtering false positives. 

%\subsection{Quatitative and Qualitative Evaluation}
It can be seen from Table~\ref{evaluation} that the overall F1-score of our proposed strategy is comparable with all the SOTA two-stage architectures. Simple multi-spectral approach is $54\times$ faster as compared to other strategies but results in many false positives and less F1-score. On the other hand our proposed approach is $21\times$ faster while retaining high F1-Score. Testing dataset images are $256\times 256$~pixels for quantitative evaluation, If the image size is larger than $256\times256$~pixels, it is detected in two image patches. To deal with this issue, we describe the duplicates in Table~\ref{evaluation}. Our proposed architecture also outperforms in region of Afghanistan where kiln and non-kiln regions exhibit extremely low contrast as illustrated in our qualitative evaluation in Fig.~\ref{fig:Var} (ii) (see training parameters, Kiln-Net vs. proposed approach and compute cost comparison in Appendix~\ref{appB}, \ref{appC} \& \ref{appD} respectively). %As seen in Fig.~\ref{fig:Var} our proposed two-stage strategy detects axis aligned kilns as compared to KilnNet architecture~\cite{nazir2020kiln}.
\vspace{-0.25cm}
\section{Conclusion and Future Work}
This paper proposes a fusion of spatio-temporal multi-spectral data with high-resolution imagery for detection of brick kilns within the ``Brick-Kiln-Belt'' of South Asia. To achieve this, we first perform classification using low-resolution spatio-temporal multi-spectral data from Sentinel-2 imagery utilizing spectral indices. Then orientation aware object detector: modified YOLOv3 (with $\theta$ value) is implemented for removal of false detections and fine-grained localization. Our proposed technique results in a $21\times$ improvement in speed with comparable or higher accuracy when tested over multiple countries. 
%In addition, we calculated heat signature values from Landsat-8 at obtained kiln locations to identify illegal kiln activity during winter smog period and compared it with data from Environmental Protection Agency. 
In future, we also aim to evaluate our proposed strategy and detection of illegal brick kiln activity during winter smog period on all over the ``Brick-Kiln-Belt'' of South Asia. Remote identification of illegal industrial activity can improve monitoring of carbon production and any forced labour in aid of law enforcement, spatial planning, and climate mitigation policy-making.
% We proposed multispectral 2 stage strategy using spatio-temporal remote sensing imagery. In $1^{st}~$step classification is performed using spectral properties and in $2^{nd}$ step orientation aware detector is applied on positive kiln detections. Our proposed strategy achieved average $0.88$ F1-score on all testing regions and is $21\times~$faster on average than the state-of-the-art two-stage architectures. Our proposed solution would enable regional monitoring and evaluation mechanisms for the Sustainable Development Goals. 
\bibliography{refs}

\begin{thebibliography}{24}
\providecommand{\natexlab}[1]{#1}
\providecommand{\url}[1]{\texttt{#1}}
\expandafter\ifx\csname urlstyle\endcsname\relax
  \providecommand{\doi}[1]{doi: #1}\else
  \providecommand{\doi}{doi: \begingroup \urlstyle{rm}\Url}\fi

\bibitem[Boyd et~al.(2018)Boyd, Jackson, Wardlaw, Foody, Marsh, and
  Bales]{boyd2018slavery}
Doreen~S Boyd, Bethany Jackson, Jessica Wardlaw, Giles~M Foody, Stuart Marsh,
  and Kevin Bales.
\newblock Slavery from space: Demonstrating the role for satellite remote
  sensing to inform evidence-based action related to {UN} {SDG} number 8.
\newblock \emph{{ISPRS} {J}ournal of {P}hotogrammetry and {R}emote {S}ensing},
  142:\penalty0 380--388, 2018.

\bibitem[Chuvieco et~al.(2002)Chuvieco, Martin, and
  Palacios]{chuvieco2002assessment}
Emilio Chuvieco, M~Pilar Martin, and A~Palacios.
\newblock Assessment of different spectral indices in the red-near-infrared
  spectral domain for burned land discrimination.
\newblock \emph{International Journal of Remote Sensing}, 23\penalty0
  (23):\penalty0 5103--5110, 2002.

\bibitem[Cotrufo et~al.(2018)Cotrufo, Sandu, Giulio~Tonolo, and
  Boccardo]{cotrufo2018building}
Silvana Cotrufo, Constantin Sandu, Fabio Giulio~Tonolo, and Piero Boccardo.
\newblock Building damage assessment scale tailored to remote sensing vertical
  imagery.
\newblock \emph{European {J}ournal of {R}emote {S}ensing}, 51\penalty0
  (1):\penalty0 991--1005, 2018.

\bibitem[Foody et~al.(2019)Foody, Ling, Boyd, Li, and Wardlaw]{foody2019earth}
Giles~M Foody, Feng Ling, Doreen~S Boyd, Xiaodong Li, and Jessica Wardlaw.
\newblock Earth observation and machine learning to meet {S}ustainable
  {D}evelopment {G}oal 8.7: Mapping sites associated with slavery from space.
\newblock \emph{{R}emote {S}ensing}, 11\penalty0 (3):\penalty0 266, 2019.

\bibitem[Gillespie et~al.(2019)Gillespie, Madson, Cusack, and
  Xue]{gillespie_changes_nodate}
Thomas Gillespie, Austin Madson, Conor Cusack, and Yongkang Xue.
\newblock Changes in {NDVI} and population in protected areas on the {Tibetan}
  plateau.
\newblock \emph{Arctic, Antarctic, and Alpine Research}, 51\penalty0
  (1):\penalty0 428--439, 2019.
\newblock \doi{http://dx.doi.org/10.1080/15230430.2019.165054}.
\newblock URL \url{https://escholarship.org/uc/item/65t7r81p}.

\bibitem[Haque et~al.(2022)Haque, Shahriar, Nahar, and Haque]{haque2022impact}
Shama~E Haque, Minhaz~M Shahriar, Nazmun Nahar, and Md~Sazzadul Haque.
\newblock Impact of brick kiln emissions on soil quality: A case study of
  ashulia brick kiln cluster, bangladesh.
\newblock \emph{Environmental Challenges}, 9:\penalty0 100640, 2022.

\bibitem[He et~al.(2016)He, Zhang, Ren, and Sun]{he2016deep}
Kaiming He, Xiangyu Zhang, Shaoqing Ren, and Jian Sun.
\newblock Deep residual learning for image recognition.
\newblock In \emph{{IEEE} {C}onference on {C}omputer {V}ision and {P}attern
  {R}ecognition}, pp.\  770--778, 2016.

\bibitem[He et~al.(2017)He, Gkioxari, Doll{\'a}r, and Girshick]{he2017mask}
Kaiming He, Georgia Gkioxari, Piotr Doll{\'a}r, and Ross Girshick.
\newblock Mask {R-CNN}.
\newblock In \emph{{IEEE} {I}nternational {C}onference on {C}omputer {V}ision},
  pp.\  2961--2969, 2017.

\bibitem[Huete et~al.(2002)Huete, Didan, Miura, Rodriguez, Gao, and
  Ferreira]{huete2002overview}
Alfredo Huete, Kamel Didan, Tomoaki Miura, E~Patricia Rodriguez, Xiang Gao, and
  Laerte~G Ferreira.
\newblock Overview of the radiometric and biophysical performance of the modis
  vegetation indices.
\newblock \emph{Remote sensing of environment}, 83\penalty0 (1-2):\penalty0
  195--213, 2002.

\bibitem[Jackson et~al.(2018)Jackson, Bales, Owen, Wardlaw, and
  Boyd]{jackson2018analysing}
Bethany Jackson, Kevin Bales, Sarah Owen, Jessica Wardlaw, and Doreen~S Boyd.
\newblock Analysing slavery through satellite technology: How remote sensing
  could revolutionise data collection to help end modern slavery.
\newblock \emph{J. {M}od. {S}lavery}, 4:\penalty0 169--199, 2018.

\bibitem[Landman \& Silverman(2019)Landman and
  Silverman]{landman2019globalization}
Todd Landman and Bernard~W Silverman.
\newblock Globalization and modern slavery.
\newblock \emph{{P}olitics and {G}overnance}, 7\penalty0 (4), 2019.

\bibitem[Li et~al.(2018)Li, Chen, Qi, Dou, Fu, and Heng]{li2018h}
Xiaomeng Li, Hao Chen, Xiaojuan Qi, Qi~Dou, Chi-Wing Fu, and Pheng-Ann Heng.
\newblock H-denseunet: hybrid densely connected unet for liver and tumor
  segmentation from ct volumes.
\newblock \emph{IEEE {T}ransactions on {M}edical {I}maging}, 37\penalty0
  (12):\penalty0 2663--2674, 2018.

\bibitem[Liu et~al.(2016)Liu, Anguelov, Erhan, Szegedy, Reed, Fu, and
  Berg]{liu2016ssd}
Wei Liu, Dragomir Anguelov, Dumitru Erhan, Christian Szegedy, Scott Reed,
  Cheng-Yang Fu, and Alexander~C Berg.
\newblock Ssd: Single shot multibox detector.
\newblock In \emph{European {C}onference on {C}omputer {V}ision}, pp.\  21--37.
  Springer, 2016.

\bibitem[Maithel(2014)]{KilnFacts}
Sameer Maithel.
\newblock Factsheets about brick kilns in south and south-east asia.
\newblock \emph{Greentech Knowledge Solutions}, 2014.

\bibitem[Misra et~al.(2019)Misra, Takeuchi, and Imasu]{misrabrick}
Prakhar Misra, Wataru Takeuchi, and Ryoichi Imasu.
\newblock Brick kiln detection in north {I}ndia with sentinel imagery using
  deep learning of small datasets.
\newblock \emph{In {A}sian {C}onference of {R}emote {S}ensing}, 2019.

\bibitem[Misra et~al.(2020)Misra, Imasu, Hayashida, Arbain, Avtar, and
  Takeuchi]{misra2020mapping}
Prakhar Misra, Ryoichi Imasu, Sachiko Hayashida, Ardhi~Adhary Arbain, Ram
  Avtar, and Wataru Takeuchi.
\newblock Mapping brick kilns to support environmental impact studies around
  delhi using sentinel-2.
\newblock \emph{ISPRS International Journal of Geo-Information}, 9\penalty0
  (9):\penalty0 544, 2020.

\bibitem[Nazir et~al.(2020)Nazir, Mian, Sohail, Taj, and Uppal]{nazir2020kiln}
Usman Nazir, Usman~Khalid Mian, Muhammad~Usman Sohail, Murtaza Taj, and Momin
  Uppal.
\newblock Kiln-net: A gated neural network for detection of brick kilns in
  south asia.
\newblock \emph{IEEE Journal of Selected Topics in Applied Earth Observations
  and Remote Sensing}, 13:\penalty0 3251--3262, 2020.

\bibitem[Ogashawara \& Bastos(2012)Ogashawara and Bastos]{Ogashawara_2012}
Igor Ogashawara and Vanessa Bastos.
\newblock A quantitative approach for analyzing the relationship between urban
  heat islands and land cover.
\newblock \emph{Remote Sensing}, 4\penalty0 (11):\penalty0 3596–3618, Nov
  2012.
\newblock ISSN 2072-4292.
\newblock \doi{10.3390/rs4113596}.
\newblock URL \url{http://dx.doi.org/10.3390/rs4113596}.

\bibitem[Ranagalage et~al.(2018)Ranagalage, Estoque, Zhang, and
  Murayama]{spatial}
Manjula Ranagalage, Ronald Estoque, Xinmin Zhang, and Yuji Murayama.
\newblock Spatial changes of urban heat island formation in the colombo
  district, sri lanka: Implications for sustainability planning.
\newblock \emph{Sustainability}, 10, 04 2018.
\newblock \doi{10.3390/su10051367}.

\bibitem[Redmon \& Farhadi(2018)Redmon and Farhadi]{redmon2018yolov3}
Joseph Redmon and Ali Farhadi.
\newblock {YOLO}v3: An incremental improvement.
\newblock \emph{arXiv preprint arXiv:1804.02767}, 2018.

\bibitem[Ren et~al.(2015)Ren, He, Girshick, and Sun]{ren2015faster}
Shaoqing Ren, Kaiming He, Ross Girshick, and Jian Sun.
\newblock Faster {R-CNN}: Towards real-time object detection with region
  proposal networks.
\newblock In \emph{Advances in {N}eural {I}nformation {P}rocessing {S}ystems},
  pp.\  91--99, 2015.

\bibitem[Xie et~al.(2016)Xie, Jean, Burke, Lobell, and Ermon]{xie2016transfer}
Michael Xie, Neal Jean, Marshall Burke, David Lobell, and Stefano Ermon.
\newblock Transfer learning from deep features for remote sensing and poverty
  mapping.
\newblock In \emph{{AAAI} {C}onference on {A}rtificial {I}ntelligence}, 2016.

\bibitem[You et~al.(2017)You, Li, Low, Lobell, and Ermon]{you2017deep}
Jiaxuan You, Xiaocheng Li, Melvin Low, David Lobell, and Stefano Ermon.
\newblock Deep gaussian process for crop yield prediction based on remote
  sensing data.
\newblock In \emph{{AAAI} {C}onference on {A}rtificial {I}ntelligence}, 2017.

\bibitem[Zhang et~al.(2019)Zhang, Wu, Xu, Wang, and Sun]{zhang2019r}
Shaoming Zhang, Ruize Wu, Kunyuan Xu, Jianmei Wang, and Weiwei Sun.
\newblock R-cnn-based ship detection from high resolution remote sensing
  imagery.
\newblock \emph{Remote Sensing}, 11\penalty0 (6):\penalty0 631, 2019.

\end{thebibliography}
\bibliographystyle{iclr2023_conference}

%\appendix
\begin{appendices}

%{\bf SUPPLEMENTARY MATERIAL}:

%\section{Appendix}
\section{Multi-spectral Indices}\label{appA}
\begin{figure}[t]
\scalebox{0.9}{
\begin{tabular}{ccc}
	\includegraphics[width=.31\columnwidth,trim={2cm 2cm 2cm 2cm}]{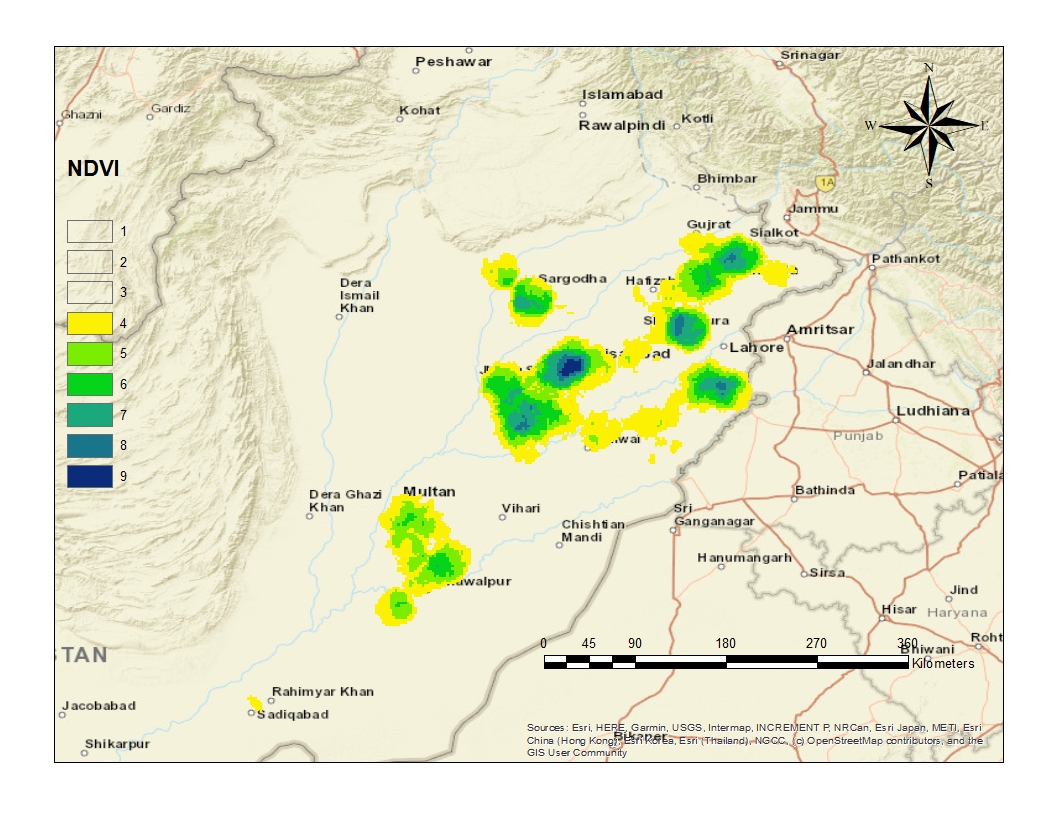}
	& \includegraphics[width=.31\columnwidth,trim={2cm 2cm 2cm 2cm}]{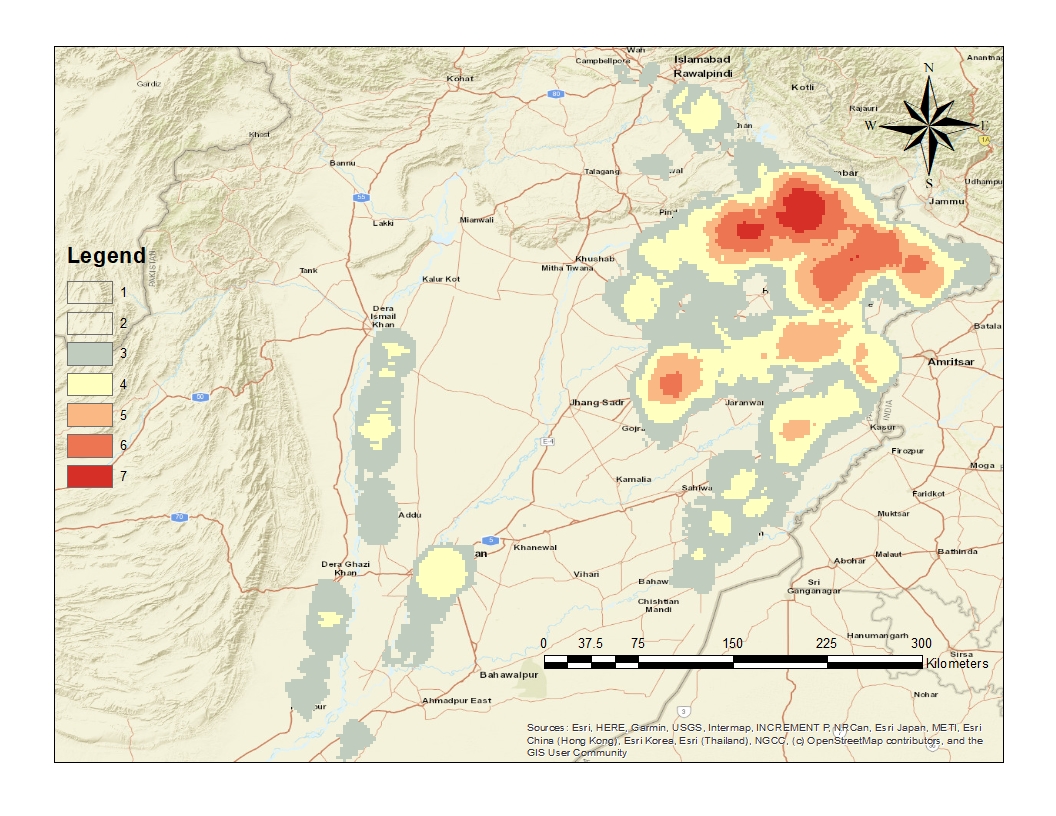}
	& \includegraphics[width=.31\columnwidth,trim={2cm 2cm 2cm 2cm}]  {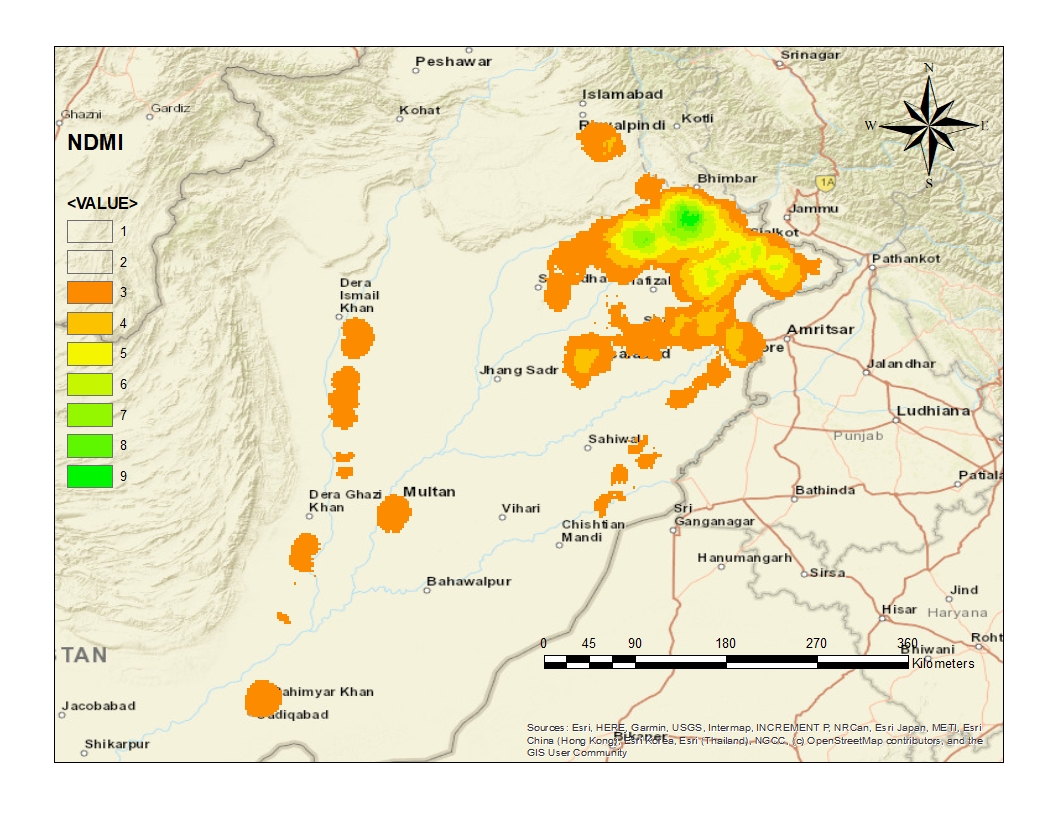} \\
	NDVI & NDBI & NDMI 
	\end{tabular}}
	\caption{Spectral Indices of kiln locations of Punjab, Pakistan (Darker colour shows more index value).}
	\label{fig:spectralIndices}
\end{figure}

We use following five spectral indices to classify potential brick kilns on Google Earth Engine:

\textbf{NDVI:}
Normalized Difference Vegetation Index (NDVI) quantifies vegetation from remote sensing imagery and  is used in various applications such as tracking population changes \cite{gillespie_changes_nodate} and spatial changes of a region \cite{spatial}. Near Infra-Red (NIR) and Red bands of remote sensing images are used by the index. NDVI always ranges from -1 to 1. The equation to find NDVI is given in Eq. \ref{eq:NDVI}. Fig.~\ref{fig:spectralIndices} shows NDVI image at brick kiln locations in region of Punjab, Pakistan and Fig.~\ref{spectra} shows the zoomed version at one kiln location. 

 \begin{equation}
\small NDVI = \frac{NIR - Red}{NIR + Red}
      \label{eq:NDVI}
  \end{equation}
  
\textbf{EVI:}
The Enhanced Vegetation Index (EVI) is designed to minimize saturation and background effects in NDVI~\cite{huete2002overview}. Since it is not a normalized difference index, compute it with this expression:
\begin{equation}
\small    EVI = \frac{2.5 * (NIR - Red)}{NIR + 6*RED -7.5*BLUE + 1}
\end{equation}

\textbf{NDBI:}
Normalized Difference Built-up Index (NDBI) is utilized to extract built-up features using remote sensing imagery. Its other applications include tracking spatial changes of a region \cite{spatial} and to find relation between urban heat islands and land cover \cite{Ogashawara_2012}. Short Wave Infra-Red (SWIR) and Near Infra-Red (NIR) bands of remote sensing images are used by the index. NDBI ranges from -1 to 1. The equation to find NDBI is given in Eq.~\ref{eq:NDBI} (see also Fig.~\ref{spectra}). 
\begin{equation}
\small      NDBI = \frac{SWIR - NIR}{SWIR + NIR}
      \label{eq:NDBI}
  \end{equation}

\textbf{NDMI:}
Normalized Difference Moisture Index (NDMI) is an index which is utilized to extricate water bodies from satellite imagery. Green and Short Wave Infra-Red (SWIR) bands of remote sensing images are used by the index. NDMI ranges from -1 to 1. The equation to find NDMI is given in Eq. \ref{eq:NDMI}. Fig.~\ref{spectra} shows NDMI image at a brick kiln location. NDMI is used to find relation between urban heat islands and land cover \cite{Ogashawara_2012} as well as several other applications.
\begin{equation}
\small NDMI = \frac{Green - SWIR}{Green + SWIR}
      \label{eq:NDMI}
  \end{equation}
  
 \textbf{BAI:}
 The Burned Area Index (BAI) was developed by \cite{chuvieco2002assessment} to assist in the delineation of burn scars and assessment of burn severity.  It is based on the spectral distance to charcoal reflectance. We used following expression to compute BAI.
 \begin{equation}
\small BAI = \frac{1.0}{(0.1 -RED)^2 + (0.06 -NIR)^2)}
 \end{equation}

\subsection{Qualitative analysis of spectral indices}\label{appA1}
\begin{figure}
    \centering
\scalebox{0.57}{
  \begin{tabular}{cccc}
            \includegraphics[width=.4\columnwidth]{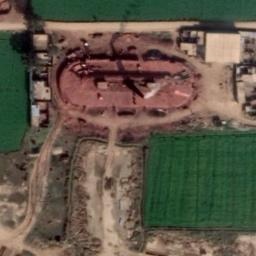} &
			\includegraphics[width=.4\columnwidth]{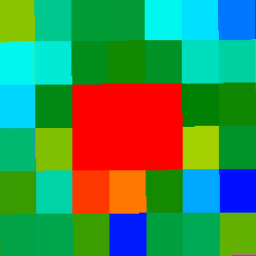} &
			\includegraphics[width=.4\columnwidth]{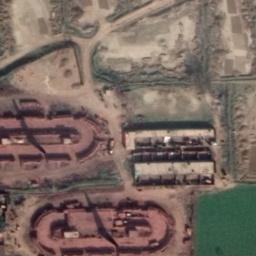} &
			\includegraphics[width=.4\columnwidth]{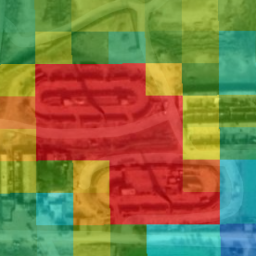} \\
			(a) & (b) & (c) & (d) \\

    \end{tabular}}
    \caption{Exemplary brick kiln images as seen in the reference high resolution ($\frac{0.5~meter}{pixel}$) imagery (a, c) [Courtesy: Digital Globe, Google Earth] and the low resolution ($\frac{10~meters}{pixel}$) Sentinel-2 imagery (b, d) with Mean NDVI values with $Opacity = 1~\&~0.5$ resp. [Image courtesy: Google Earth Engine].}
    \label{brickkilnpixels}
\end{figure}

\begin{figure}[h]
\centering
\scalebox{0.95}{
\begin{tabular}{ccccc}
	\includegraphics[width=.15\columnwidth]{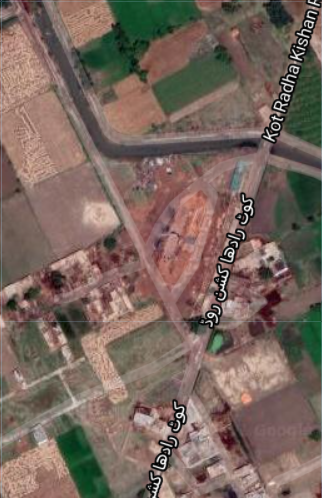}&
	\includegraphics[width=.15\columnwidth]{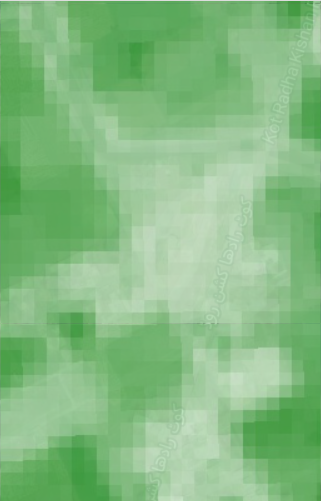}
	& \includegraphics[width=.15\columnwidth]{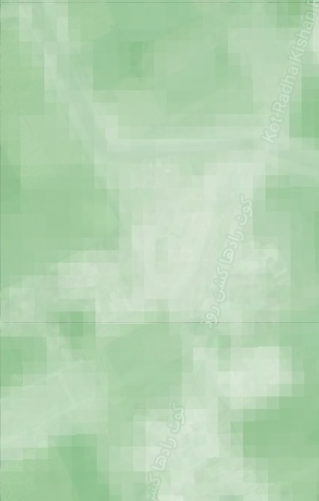}
	& \includegraphics[width=.15\columnwidth]{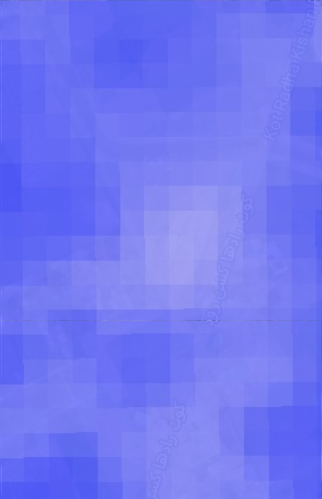}&
	\includegraphics[width=.15\columnwidth]{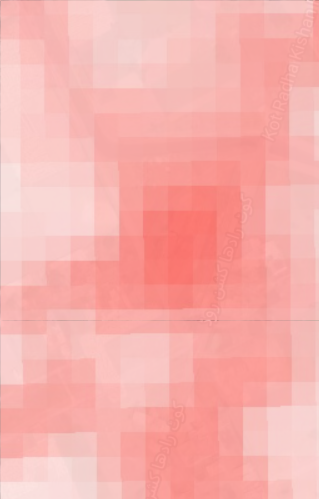}
	\\
	Satellite Image & EVI & NDVI & NDMI & NDBI  
	\end{tabular}}
	\caption{\small Kiln surrounding has a low vegetation index (EVI, NDVI), low moisture Index (NDMI) and a high built-up index (NDBI); (Darker colour shows more index value)~[Image courtesy: Google Earth Engine].}
	\label{spectra}
\end{figure}

The kiln surrounding has a low vegetation index (EVI, NDVI), low moisture index (NDMI) and a high built-up index (NDBI) (see Fig.~\ref{brickkilnpixels} and Fig~\ref{spectra}). Thus in this work we classify brick kilns using mixture of spectral indices namely Normalized Difference Vegetation Index (NDVI), Enhanced Vegetation Index (EVI), Normalized Difference Built-up Index (NDBI), Normalized Difference Moisture Index (NDMI) and Burned Area Index (BAI).

\section{Hyperparameters for the training of Orientation-aware YOLOv3}\label{appB}
Optimization method is Adam with an initial learning rate of $0.001$. The learning rate increases by $0.1$ if validation loss does not decline for $20$ epochs. Instead of using fixed number of epochs, we used early stopping criteria which terminates the training process in case there is no improvement for $50$ consecutive epochs.Unlike~\cite{zhang2019r}, in our work instead of regressing out the value of $\theta$ directly, the methodology of Vanilla YOLOv3~\cite{redmon2018yolov3} was retained but instead of using only one class for the detection of brick kilns, the following $5$ classes were used: Kiln-$0^o$, Kiln-$20^o$, Kiln-$40^o$, Kiln-$140^o$, and Kiln-$160^o$ which rotated the un-oriented bounding box (predicted by vanilla YOLOv3) by $0$, $20$, $40$, $140$ and $160$ degrees respectively based on their orientation. Using quantized values of $\theta$ rather than regression reduces the search space and time taken in the training and testing/prediction stages.
\section{KilnNet~\cite{nazir2020kiln} vs. Proposed Approach}\label{appC}
The Key differences between KilnNet~\cite{nazir2020kiln} and our proposed approach are as follows (see Fig.~\ref{fig:Var}):
\begin{itemize}
\item In the prior approach~\cite{nazir2020kiln}, high resolution DigitalGlobe imagery is used. Although use of high-resolution imagery improves the accuracy, as mentioned in Fig. 10 of \cite{nazir2020kiln}, it requires 93 days to process the entire brick kiln belt ($1551997~ km^2$) of South Asia. In this work we addressed this concern and proposed an improved approach that results in a $21\times$ improvement in
speed with comparable or higher accuracy when tested over multiple
countries. We achieved this by a strategy that uses low-resolution multi-spectral imagery as well as high-resolution imagery. The low-resolution multi-spectral pre-filter bulk of data without introducing missed detections. Thus the detailed analysis is only applied on a very small chunk of high-resolution imagery.
\item In the prior methodology, two-stage gated neural network architecture consisting of a ResNet-152 classifier and a YOLOv3 detector is proposed. Our proposed coarse-to-fine strategy is also a two-staged approach with two major differences. In the first stage we replace CNN based classification on high-resolution ($0.149~\frac{meter}{pixel}$) data with fusion of low-resolution ($10~\frac{meter}{pixel}$) spectral indices to pre-filter bulk of the date followed by orientation-aware YOLOv3 detector on the filtered data. Furthermore, in this work we proposed a modified object detector that along with bounding boxes also produces the orientation information. 
%which resulting ##% improvement in localization accuracy.
\item In the previous methodology, classifier is selected by considering the class imbalance issues using F-Beta measure. In the proposed approach we select a classifier based on the observation that kiln locations have low vegetation and moisture index whereas high build-up and burn area index.
\item The prior methodology takes $674$ seconds to process three datasets in Table~\ref{evaluation} of the paper. As a result of the above mentioned differences the proposed approach on the other hand reduces this compute time to $38$ seconds only.
\end{itemize}

In nutshell, our coarse-to-fine strategy aims to filter the bulk of the data using spectral properties on Google Earth Engine while the detector is only applied on small amount of positive detections as to generate localization information while filtering false positives on Google Colab (Tesla T4). We evaluated our trained network on unseen dataset consisting of Kasur (Pakistan), New Delhi (India) and Deh Sabz (Afghanistan) for quantitative analysis. (see Table ~\ref{evaluation}. If the brick kiln is larger than 256 x 256 it is detected in two image patches. To deal with this issue we describes the duplicates in Table~\ref{evaluation}.

\section{Compute Cost Comparison with State-of-the-Art}\label{appD}
The detection of brick kilns by analyzing the spectral properties on Google Earth Engine takes approximately $3$, $4$ and $8$ seconds to process Kasur (Pakistan), New Delhi (India) and Deh Sabz (Afghanistan), respectively. Then the orientation aware detector: YOLOv3 is ran on potential brick kiln locations including false positives on Google Colab Tesla T4 GPU. Each experiment is repeated 5 times to find the the average time on Google Colab which is 4.04, 4.16 and 15.1 to localize potential kilns in Kasur (Pakistan), New Delhi (India) and Deh Sabz (Afghanistan), respectively (see Table~\ref{evaluation}).
\end{appendices}

%\subsection{Key differences between proposed approach and %KilnNet~\cite{nazir2020kiln}}

%
% =========================================================
%
%\subsection{Detection of Illegal Kiln Activity}
%After identifying the kiln locations we analyzed the heat signature values on these locations using Landsat-8 Satellite's Thermal Infrared 1 Band (band 10) at Tehsil level. We also collected the ground truth values: 1) No. of illegal kiln operations during smog period 2) No. of FIR lodged against brick kiln' owner and number of sealed brick kilns from Environmental Protection Authority (EPA). We find out that the heat signature value at tehsil level directly correlates with the number of illegal kiln operations in that tehsil (see Fig.~\ref{IllegalActivity}).
% We presented detailed experimentation of our proposed multispectral two-stage strategy with comparison with state-of-art classifiers: ResNet-152, Inception-ResNet-v2, Inception-v3 and detectors: Faster R-CNN , SSD and YOLOv3.

\end{document}